\documentclass{article}
\usepackage{spconf,amsmath,graphicx}
\usepackage{hyperref}

\usepackage{enumitem}
\setlist{nosep, leftmargin=14pt}

\usepackage{mwe} 

\title{SarcNet: A Novel AI-based Framework to Automatically Analyze and Score Sarcomere Organizations in Fluorescently Tagged hiPSC-CMs}
\name{Huyen Le$^{1,2}$ \qquad Khiet Dang$^{5}$ \qquad Tien Lai$^{2,4}$ \qquad  Nhung Nguyen$^{3}$ \qquad  Mai Tran$^{1,2}$ \qquad Hieu Pham$^{1,2,*}$\thanks{* Corresponding author: hieu.ph@vinuni.edu.vn (Hieu Pham)}}
\address{   $^{1}$ College of Engineering \& Computer Science, VinUniversity \\
            $^{2}$ VinUni-Illinois Smart Health Center, VinUniversity \\
         $^{3}$ College of Health Science, VinUniversity \\
         $^{4}$ Department of Electrical \& Computer Engineering, University of Illinois Urbana-Champaign (UIUC)\\
         $^{5}$ Faculty of Sciences and Technologies, University of Burgundy}

\begin{document}
%
\maketitle
\begin{abstract}
Quantifying sarcomere structure organization in human-induced pluripotent stem cell-derived cardiomyocytes (hiPSC-CMs) is crucial for understanding cardiac disease pathology, improving drug screening, and advancing regenerative medicine. Traditional methods, such as manual annotation and Fourier transform analysis, are labor-intensive, error-prone, and lack high-throughput capabilities. This paper presents a novel deep learning-based framework that leverages cell images and integrates cell features to automatically evaluate the sarcomere structure of hiPSC-CMs from the onset of differentiation. The proposed framework contains the SarcNet, a cell-features concatenated and linear layers-added ResNet-18 module, to output a continuous score ranging from 1 to 5 that captures the level of sarcomere structural organization. SarcNet achieves a Spearman correlation of 0.831 with expert evaluations, demonstrating superior performance and an improvement of 7.5\% over the previous approach, which uses linear regression. Our results also show a consistent pattern of increasing organization from day 18 to day 32 of differentiation, aligning with expert evaluations. This study contributes to the development of a new automated tool for quantifying sarcomere structural organization in hiPSC-CMs, advancing cardiac research.
\end{abstract}
\begin{keywords}
hiPSC-CMs, sarcomere, deep learning.
\end{keywords}
\section{Introduction}
\label{sec:intro}

Human-induced pluripotent stem cell-derived cardiomyocytes (hiPSC-CMs) are cells safely collected from the blood or skin of living humans and then reprogrammed into human heart muscle cells \cite{hamledari2022using}. Fully characterized hiPSC-CMs have a range of critical applications, including disease modeling, drug discovery, and precision medicine \cite{bedada2016maturation}. Specifically, these cells have attracted considerable attention as a promising alternative for modeling arrhythmogenic disorders and assessing the cardiac function of patients \cite{yang2022use}. In another instance, Vicente \textit{et al.} \cite{vicente2018mechanistic} introduced the Comprehensive in vitro proarrhythmia Assay (CiPA) as a method to assess the impact of drugs on various ion channels in hiPSC-CMs and to predict the risk of proarrhythmia.

To effectively utilize these applications, hiPSC-CMs must accurately reflect the electrical activity, calcium dynamics, structure, and contractility of mature cardiomyocytes \cite{bedada2016maturation}. Recently, various methods have been employed to assess the maturation of hiPSC-CMs, using a diversity of characteristics, such as sarcomere configuration, electrophysiological attributes, metabolism, and gene expression profiles \cite{ahmed2020brief, bedada2014acquisition}. Among them, evaluating the structure of the sarcomere, the basic contractile unit of hiPSC-CMs, based on imaging data is an important technique to consider. The reason is that the organization and structure of the sarcomeres and myofilaments are closely tied to the contraction capabilities of adult cardiomyocytes \cite{skorska2022monitoring}. In particular, mature hiPSC-CMs are notably longer and display a higher degree of structural organization compared to their immature counterparts \cite{machiraju2019current}. 

Several studies have employed artificial intelligence (AI) to quantify maturity of sarcomere structure in hiPSC-CMs by analyzing fluorescent images \cite{pasqualini2015structural,gerbin2021cell}.  Pasqualini \textit{et al.} \cite{pasqualini2015structural} first defined a set of 11 metrics to capture the increasing organization of sarcomeres within striated muscle cells during their developmental process and then used machine learning algorithms to score the phenotypic maturity of hiPSC-CMs unbiasedly. However, due to the limited availability of hiPSC-CMs at various developmental stages, the model was trained on primary cardiomyocytes from neonate rats (rpCMs) and was tested on immature hiPSC-CMs only. Gerbin \textit{et al.} \cite{gerbin2021cell} used linear regression with 11 cell features to distinguish stages of myofibrillar organization in hiPSC-CMs. The model reached a Spearman correlation of 0.63 and 0.67 on two testing sets, respectively. Despite these advances, further development is needed since clinical applications require higher performance.

To address the gap, we propose in this study SarcNet, a cell-features concatenated and linear layers-added ResNet-18 convolutional neural network (CNN) \cite{he2016deep} for automatically quantifying sarcomere structural organizations on single-cell images of hiPSC-CMs. In particular, the proposed model leverages predictions by concatenating the output from the ResNet-18 module and a representation vector of quantitative single-cell measurements of subcellular organization. In addition, we assess the performance of the proposed framework using two different feature extraction protocols. We further assess the explainability of the proposed model by utilizing Gradient-weighted Class Activation Mapping (Grad-CAM) \cite{selvaraju2017grad}. Our main contributions can be summarized as follows:
\begin{enumerate}
\item We introduce a novel AI-based framework to effectively score sarcomere organizations in fluorescently tagged hiPSC-CMs single cell images by integrating feature extraction with the state-of-the-art CNN model.
\item We conduct thorough experiments to evaluate the efficacy of the proposed solution. The experimental findings demonstrated that SarcNet model enhances performance by 7.5\% for Spearman correlation metric compared to linear regression, the previously used method. Notably, we can speed up the training and inference by simplifying the feature extraction process while maintaining performance.
\item Our codes are released at \href{https://github.com/vinuni-vishc/sarcnet}{https://github.com/vinuni-vishc/sarcnet}.
\end{enumerate}
The rest of the paper covers the proposed approach (Section \ref{methodology}), the experiments and results (Section \ref{Experiments}), and concludes with Section \ref{Concludes}. 
\section{METHODOLOGY}
\label{methodology}
\label{sec:format}

\subsection{Problem Formulation} \label{Problem_formulation}
In a regression problem setting, we are given a training set \( \mathcal{D} \) consisting of \textit{N} samples, \( \mathcal{D} \) = \{\(( x^{(i)}, y^{(i)} \)); (i) = 1,..., \textit{N}\} where each input image \(x^{(i)} \in \mathcal{X}\) is associated with a continuous value \(y^{(i)} \in \mathcal{Y}\). Our goal is to approximate a mapping function \(f_{(\theta)} : \mathcal{X} \rightarrow \mathcal{Y}\) from input images, which are fluorescently tagged single-cell images of hiPSC-CMs to continuous output variables representing its sarcomere structural organization. This learning task could be performed by training a CNN that minimize the MSE loss function, which is given by
\begin{equation}
   \mathcal{L}(\theta) = \frac{1}{N} \sum_{i=1}^{N} |y^{(i)} - \hat{y}^{(i)}|^2,
\end{equation}
where \(y^{(i)}\) is the ground truth and \(\hat{y}^{(i)}\) is the predicted value.

\subsection{Overall Framework} \label{overall_framework}

\begin{figure*}[ht]
    \centering
    \includegraphics[width=0.87\textwidth]{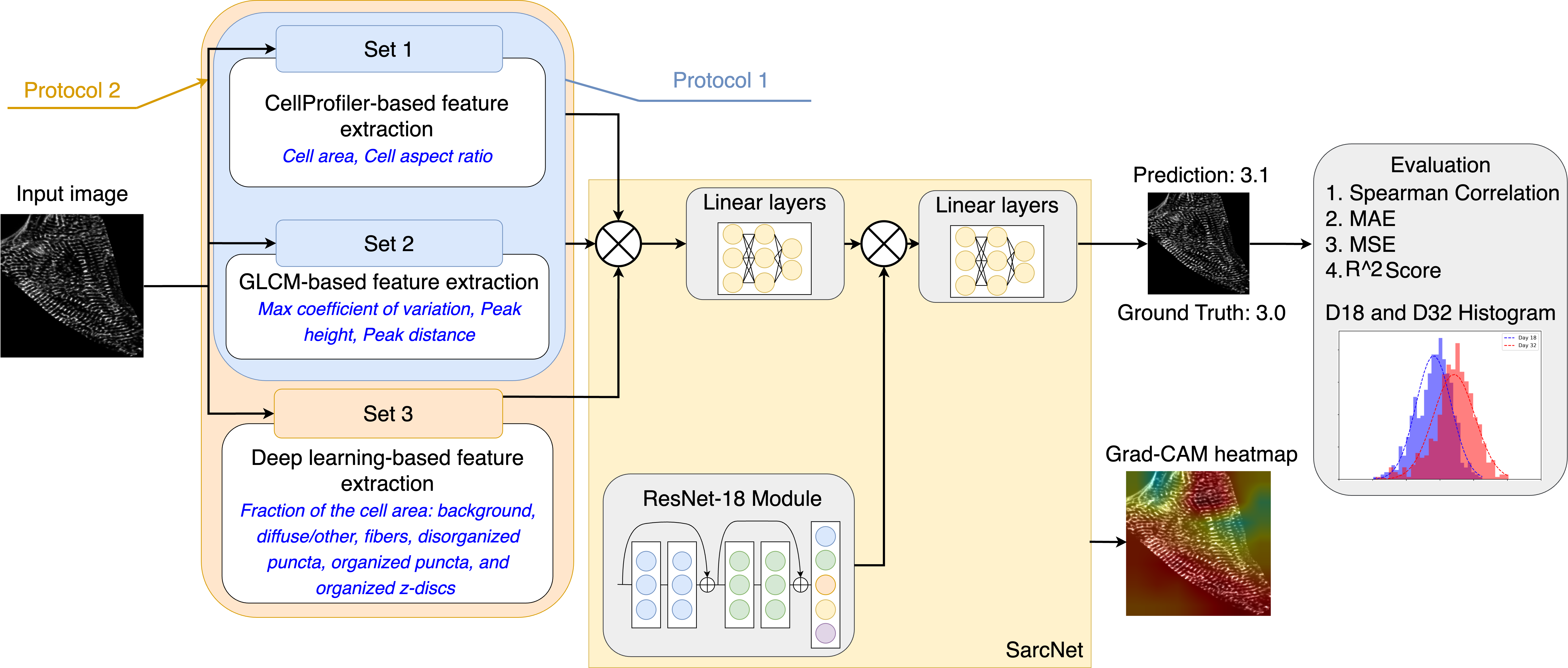}
    \caption{An illustration of the proposed overall framework, which aims to quantify sarcomere structure organization. The system takes the hiPSC-CM images as input and outputs the scores of alpha-actinin-2 patterns, ranging from 1 to 5. Grad-CAM heatmaps is used to highlight important visual features.}
    \label{fig_framework}
\end{figure*}

Fig.~\ref{fig_framework} provides an overview of the proposed approach, which is a regression framework using CNN. It predicts a continuous score of sarcomere structure organization by taking fluorescent images of hiPSC-CMs single cells as input. We enhance the prediction performance by integrating feature extraction into the training procedure. Model performance is evaluated using four metrics: Spearman correlation, mean absolute error (MAE), mean squared error (MSE), and $R^2$ score. A visual explanation module based on Grad-CAMs is also used for model interpretation. We also examine the changes in cell organization over time and confirm the increasing maturity of sarcomere structures between day 18 and day 32 time points.
\subsection{Framework Architecture} \label{framework_archi}

\subsubsection{Feature Extraction} \label{sec:Feature extraction}
Three sets with a total of 11 cell features are extracted for each input image using techniques provided in \cite{gerbin2021cell}. More details, Set 1, the CellProfiler-based \cite{stirling2021cellprofiler} feature extraction generates two cell morphology measurement metrics: cell area and aspect ratio. Set 2, the GLCM-based \cite{haralick1973textural} method produces three metrics describing sarcomere alignment: max coefficient of variation, peak height, and peak distance. Finally, Set 3, the deep learning-based method provides six additional metrics representing the fraction of the cell area covered by each class: background, diffuse/other, fibers, disorganized puncta, organized puncta, and organized z-discs.
The features extracted from Set 1 and Set 2 together are referred to as Protocol 1. The features extracted from Set 1, Set 2, and Set 3 together are referred to as Protocol 2.

\subsubsection{ResNet-18 Module} 
ResNet-18 \cite{he2016deep}, one of the state-of-the-art image classification models, is adapted for the sarcomere structural level prediction task as the deep residual learning framework makes it well-suited for capturing the intricate hierarchical patterns of sarcomere structures. The ResNet-18 module architecture used in this study consists of 18 convolution layers, an average pooling layer, and three linear layers. Compared to the original architecture, two fully connected layers are added at the end before outputting the result.

\subsubsection{SarcNet}

The proposed SarcNet framework comprises two branches of the ResNet-18 module and linear layers as depicted in Fig.~\ref{fig_framework}. First, the former requires an input image with a size of $3 \times 224 \times 224$ to pass through the ResNet-18 module (described in \ref{framework_archi}). Second, the input feature vector would be passed through three linear layers with rectified linear unit (ReLU) activation functions to extract more deep linear representations. After concatenating with the output from the ResNet-18 module, a new representation vector is created, consisting of the information contributed equally from both input images and input features. This vector is then processed through four linear layers to get the final score.

\section{Experiments}
\label{Experiments}
\subsection{Dataset and Experimental Settings} \label{sec:Dataset and settings}
\subsubsection{Dataset}
In this study, we use a publicly available dataset of alpha-actinin-2-mEGFP-tagged hPSC-CMs single cells at days 18 and 32 time points since the initiation of differentiation, provided by the Allen Institute for Cell Science (AICS) \cite{gerbin2021cell}. The dataset consists of two components: images of single cells and tabular data containing cell features, along with the alpha-actinin-2 organization score corresponding to each image.

As described in the original paper \cite{gerbin2021cell}, each cell is manually scored for the structural maturity of its sarcomere organization by two experts. Based on the majority of their organization, experts categorize cells into five score groups, from 1 to 5. A score of 1 indicates cells with sparse, disorganized puncta; a score of 2 indicates cells with denser, more organized puncta; a score of 3 indicates cells with a combination of puncta and other structures like fibers and z-discs; a score of 4 indicates cells with regular but not aligned z-discs and finally a score of 5 indicates cells with almost organized and aligned z-discs. Cells lacking alpha-actinin-2 mEGFP protein expression were assigned a score of ``0'' and subsequently excluded from further analysis. After filtering, there are a total of 5,761 images of different sizes. Expert 1 categorized 293 images as Score 1, 708 as Score 2, 3,868 as Score 3, 798 as Score 4, and 94 as Score 5. Expert 2 classified 115 images as Score 1, 1,007 as Score 2, 2,979 as Score 3, 1,527 as Score 4, and 113 as Score 5. For consistency, we define ground truth in downstream analysis as the average score from two experts.

The tabular data component provides cell features extracted by CellProfiler, GLCM, and deep-learning model. 11 features from the tabular data in the dataset package are utilized, including cell area, cell aspect ratio, max coefficient of variation, peak height, peak distance, fraction of cell area for background, diffuse/other, fibers, disorganized puncta, organized puncta, and organized z-discs. These features are applied into two protocols as depicted in Section \ref{sec:Feature extraction}.

\subsubsection{Experimental Settings}
To evaluate the effectiveness of the proposed method, several experiments have been conducted. We divide the dataset into training, validation, and testing sets with 3,686 cells for training, 922 cells for validation, and 1,153 cells for testing. For all experiments, we preprocess the images by resizing them to a size of $3 \times 224 \times 224$ to align with the ResNet-18 input shape, followed by normalization; and preprocess the tabular data by applying standard scaler from the Scikit-learn library. To investigate the impact of feature extraction, we perform the proposed method on both Protocol 1 and Protocol 2.

\subsection{Training Methodology}
In this study, an Adam optimizer was used with a learning rate of 0.0005 to optimize the MSE objective function (Section \ref{Problem_formulation}). The model was trained with a batch size of 40 in 100 epochs using the Pytorch framework. The network with the highest Spearman correlation on the validation dataset was selected and continuously tested on the testing dataset. All models were trained on a GeForce RTX 3090 GPU.

\subsection{Experimental Results}
\label{Experimental Results}
To quantify the effectiveness of our model, we compares the performance of SarcNet with linear regression used in \cite{gerbin2021cell} and other state-of-the-art CNN models on the test set. Table~\ref{table_result} shows that SarcNet using both feature extraction protocols outperform other models, with a 7.5\%, 0.063, 0.075, and 15.7\% improvement compared to linear regression in Spearman correlation, MAE, MSE, and $R^2$ score, respectively. It is noticeable that the ResNet-18 model and SarcNet achieve an approximately similar Spearman correlation of 0.829 and 0.831, but are significantly different in the results of the other metrics. An explanation for this finding could be that the Spearman correlation only measures the association based on ranks between the GT and predictions but does not consider the distance between them as MAE and MSE. The ResNet-18 model seems to quantify the sarcomere structure organization in the correct rank of scores but still achieves poor performance in MAE and MSE. In contrast, thanks to feature extraction, SarcNet can get more relevant information and score the sarcomere structure with smaller values in MAE and MSE. The visualizations of predicted results by SarcNet are shown in Fig.~\ref{fig_gradcam}A. Notably, in both feature extraction protocols, the SarcNet model reveals nearly identical results. This results indicate that the Protocol 1, which is more simple, also contributes significantly to the overall performance. The training and inference processes therefore can be speed up by simplifying feature extraction protocol.

\begin{table}[]
\small
\centering
\caption{Quantitative assessment of SarcNet and other models on the testing dataset.}
\begin{tabular}{|c|c|c|c|c|}
\hline
Model                                                                    & \begin{tabular}[c]{@{}c@{}}Spearman \\ Corr\end{tabular} & MAE            & MSE            & \begin{tabular}[c]{@{}c@{}}$R^2$ \\ Score\end{tabular} \\ \hline
Linear Regression                                                        & 0.756                                                    & 0.373          & 0.234          & 0.514                                                                 \\ \hline
ResNet-18                                                                & 0.829                                                    & 0.453          & 0.300          & 0.379                                                                 \\ \hline
DenseNet                                                                 & 0.813                                                    & 0.340          & 0.194          & 0.599                                                                 \\ \hline
AlexNet                                                                  & 0.788                                                    & 0.346          & 0.195          & 0.596                                                                 \\ \hline
\textbf{SarcNet (Protocol 1)} & 0.825                                                    & \textbf{0.310} & \textbf{0.159} & \textbf{0.671}                                                        \\ \hline
\textbf{SarcNet (Protocol 2)} & \textbf{0.831}                                           & \textbf{0.310} & 0.161          & 0.668                                                                 \\ \hline
\end{tabular}
\label{table_result}
\end{table}

\section{Discussion and Conclusion}
\label{Concludes}
\begin{figure}
    \centering
    \includegraphics[width=8cm]{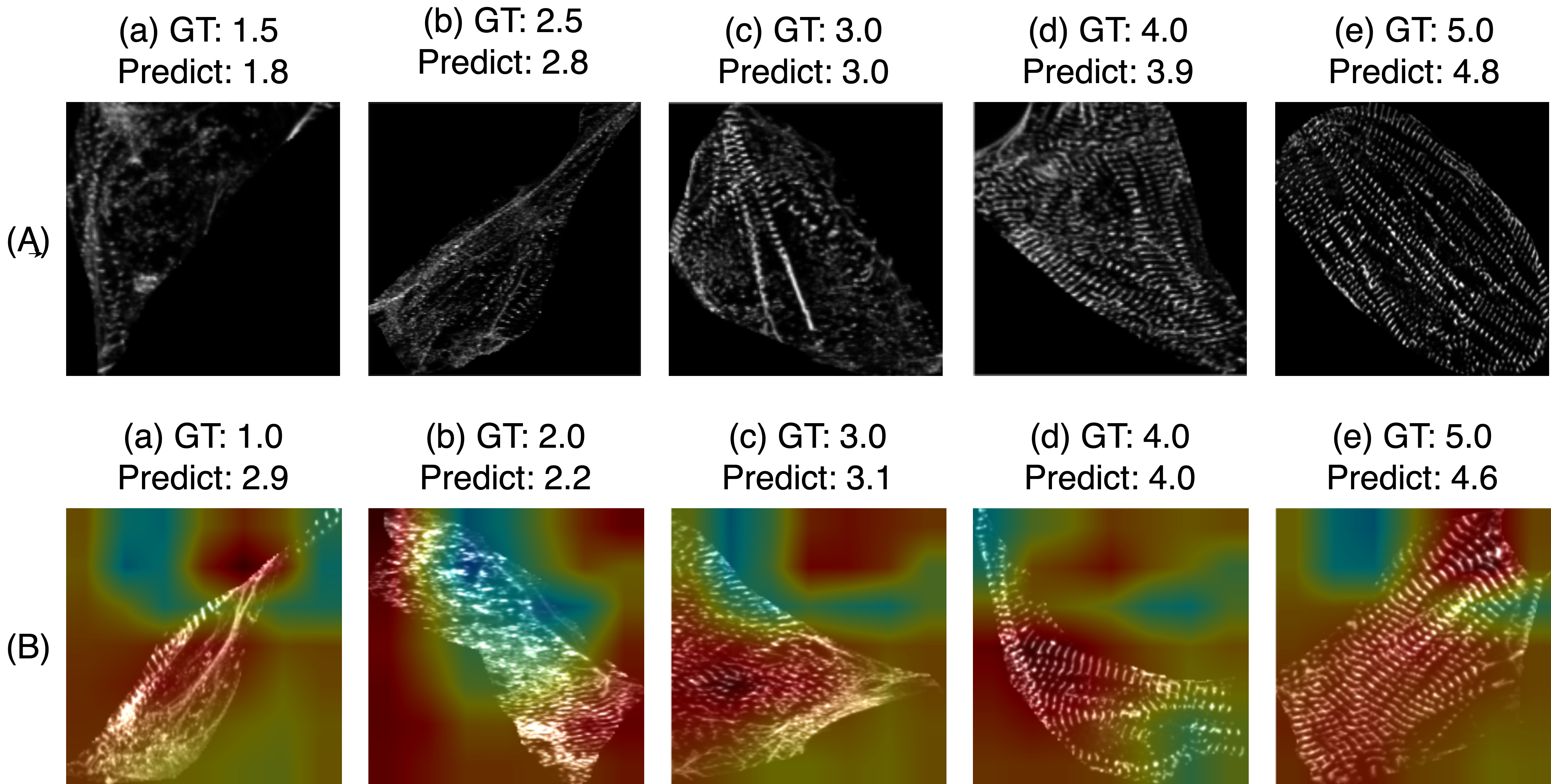}
    \caption{Visualizations of examples of prediction result (A) and Grad-CAM heatmaps (B).}
    \label{fig_gradcam}
\end{figure}

Grad-CAM heatmaps (Fig.~\ref{fig_gradcam}B) are used to interpret the SarcNet model predictions. Fig.~\ref{fig_gradcam}B(c), (d), and (e) produce the highest values focusing mainly on the sarcomere structure, while Fig.~\ref{fig_gradcam}B(b) has the lowest value at the center region of the cell. An explanation for this could be that the two sides of the cell have clearer fiber patterns while the center does not; as a result, it would be easier for the model to make decisions based on the two sides of the cell. However, Fig.~\ref{fig_gradcam}B(a) illustrates a mistaken case in which the model seems to focus more on the regions with organized z-discs than diffuse regions.

\begin{figure}
    \centering
    \includegraphics[width=7cm]{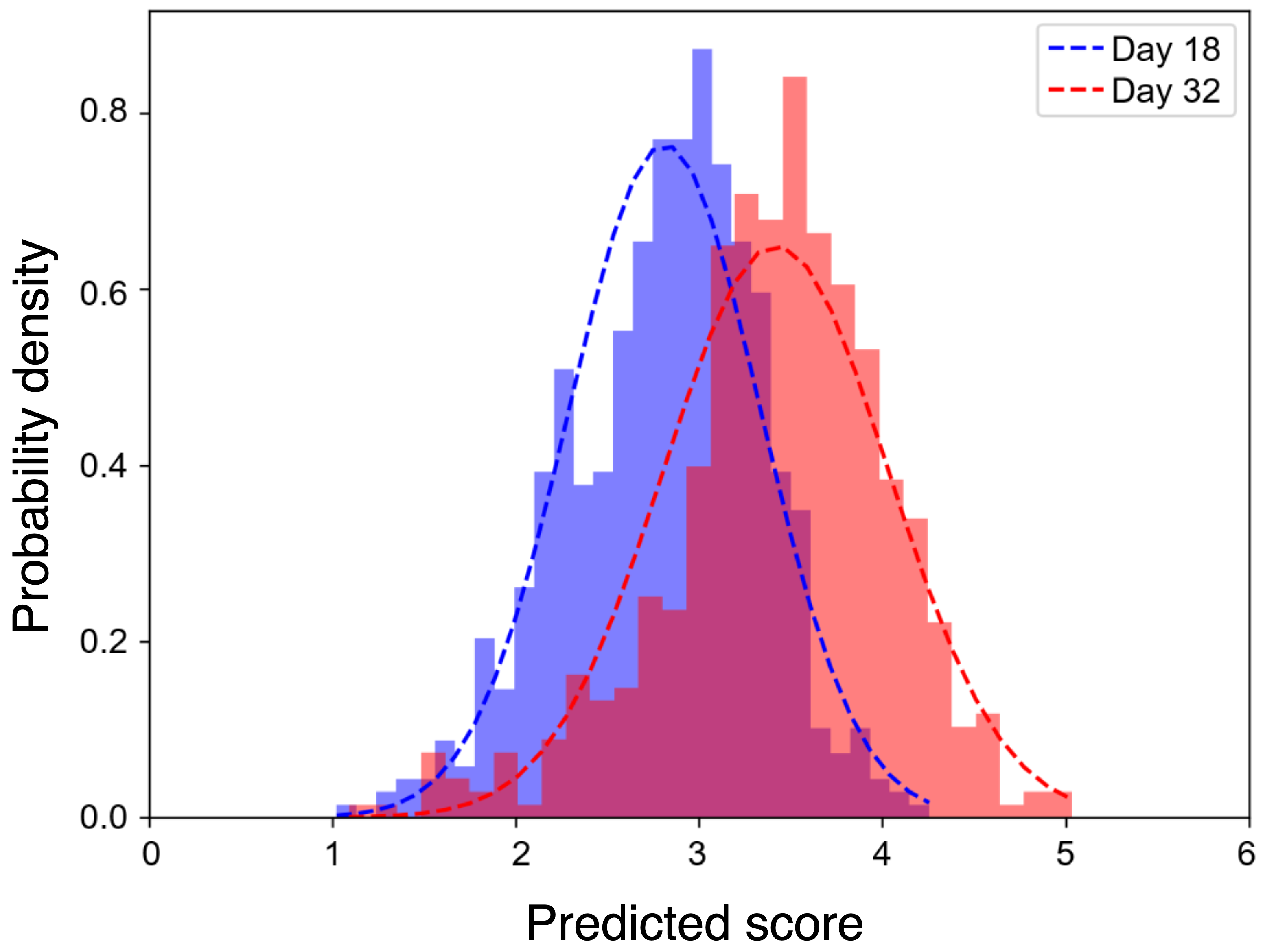}
    \caption{Histogram of the predicted scores for day 18 (blue) and day 32 (red).}
    \label{fig_cellage}
\end{figure}

Regarding changes in cell organization over time, Fig.~\ref{fig_cellage} aims to confirm the pattern of increasing organization levels between days 18 and 32. It is noticeable that the predicted scores from the SarcNet model are consistent with the culture period, with a slight shift to the right of the 32-day histogram. While most of the hiPSC-CM organization on day 18 concentrates around level 3, with most of the regions being disorganized puncta, day 32 indicates more organized regions with level 4. However, most of the cells on day 32 are still immature and need further culture. This finding agrees with the research in \cite{ebert2019proteasome}, which proclaimed that cell contractility reaches the maximum level at the 200-day time point.

To conclude, we developed SarcNet, a deep learning-based method for quantifying sarcomere structure organization. Experiments show that SarcNet outperforms other approaches, with the best performance in Spearman correlation of 0.831. It is also noticeable that we can reduce the number of features while retaining the performance. Some limitations need addressing before quantifying sarcomere structure organization at single cell level, with one of the most noteworthy is the lack of single-cell segmentation progress. Experts manually mark single-cell boundaries in the AISC dataset \cite{gerbin2021cell}, as there is no accessible automated framework. This work is challenging because the individual cells are not completely separated from one another but overlap. Future work aims to develop an automated procedure that addresses this challenge and facilitates more accurate quantification of sarcomere structure organization.

\newpage 

\section{Compliance with ethical standards}
This research study was conducted retrospectively using human subject data made available in open access by \cite{gerbin2021cell}. Ethical approval was not required as confirmed by the license attached with the open access data.

\bibliographystyle{IEEEbib}
\bibliography{refs}

\end{document}